# Model-Free $\delta$-Policy Iteration Based on Damped Newton Method for Nonlinear Continuous-Time $H_\infty$ Tracking Control

Qi Wang

*Abstract*—This paper presents a $\delta$-PI algorithm which is based on damped Newton method for the $H_\infty$ tracking control problem of unknown continuous-time nonlinear system. A discounted performance function and an augmented system are used to get the tracking Hamilton-Jacobi-Isaac (HJI) equation. Tracking HJI equation is a nonlinear partial differential equation, traditional reinforcement learning methods for solving the tracking HJI equation are mostly based on the Newton method, which usually only satisfies local convergence and needs a good initial guess. Based upon the damped Newton iteration operator equation, a generalized tracking Bellman equation is derived firstly. The $\delta$-PI algorithm can seek the optimal solution of the tracking HJI equation by iteratively solving the generalized tracking Bellman equation. On-policy learning and off-policy learning $\delta$-PI reinforcement learning methods are provided, respectively. Off-policy version $\delta$-PI algorithm is a model-free algorithm which can be performed without making use of a priori knowledge of the system dynamics. NN-based implementation scheme for the off-policy $\delta$-PI algorithms is shown. The suitability of the model-free $\delta$-PI algorithm is illustrated with a nonlinear system simulation.

## I. INTRODUCTION

THE $H_\infty$ optimal control theory is a well-known robust control method which has been widely applied in controller design for systems with disturbance, and the adverse impact of disturbance signals on controller performance can be attenuated. The $H_\infty$ control method aims to find a control policy that the closed-loop system satisfies disturbance attenuation condition, and the system dynamics is locally asymptotically stable while there is no disturbance signal [1]. The sufficient condition for the solution of the $H_\infty$ control problem is to solve the Hamilton-Jacobi-Isaacs (HJI) equation, which is a nonlinear partial differential equation [2]. For tracking problems, traditional methods to $H_\infty$ controller design involve two steps, first, design a feedforward controller to perform tracking, then, design an $H_\infty$ feedback controller to stabilize the error dynamics of the tracking system [3]-[6]. These approaches were suboptimal because the control cost generated by feedforward control inputs is ignored in the design of $H_\infty$ controller.

In order to solve the problem of neglecting the control cost caused by feedforward control component in the performance index functional, a new discounted performance functional was proposed for the tracking problem $H_\infty$ control, and a more universal definition of the $L_2$-gain was developed, where the energies of the tracking error and the whole control input were weighted by a discount factor of exponential decay type in the performance function [6]. This method corrected the traditional method of only including the control cost of the feedback control component in the performance function, and then a new tracking HJI equation concerned with the exponential discounted performance function was presented, which can give both the feedback component and feedforward component of the control input synchronously.

HJI equation is a nonlinear partial differential equation, which is difficult to obtain the closed-form solution for nonlinear systems, and the complete knowledge of the nonlinear system dynamics is required for solving the analytical solutions. Therefore, it is even worse difficult to solve the analytical solution of $H_\infty$ control in practical application scenarios, when the accurate system models cannot be obtained or acquisition cost is high. To overcome these difficulties, reinforcement learning (RL) and adaptive dynamic programming (ADP) techniques have been widely used in solving $H_2$ optimal control and $H_\infty$ optimal control problems in the past few decades, and have been successfully applied in many practical scenarios. As an interdisciplinary subject in the communities of artificial intelligence and control, RL and ADP which can be used for approximately solving the optimal control problems, were widely introduced and developed in the computational intelligence and machine learning fields [7], [8]. The solution of the HJB equation or HJI equation is not solved directly in the ADP algorithms, instead, random or directional search is made in the strategy space of admissible control. The solution includes a technology called policy iteration (PI), which was first presented in the solution framework of Markov decision problems [9]. In fact, the PI algorithm requires two iterations: policy evaluation is the first step, in this step, the value function related to the stable control strategy is evaluated and calculated; policy update is the second step, in which control strategy is improved through optimizing the value function.

Repeat these two steps until the policy update step no longer achieves optimization results, indicating that the control strategy converges to the optimal one [10]. PI algorithm is in essence the Newton's method [11]. For continuous time linear dynamic system and nonlinear dynamic system, the PI algorithm has been widely used to solve the optimal state-feedback control and differential games in [12]-[15]. Synchronous policy iteration method was introduced in [16], which implemented simultaneous continuous-time adjustment of both actor and critic neural networks, and a persistence of excitation condition was involved to guarantee convergence. However, all of the above-mentioned algorithms are model-based approaches, the complete knowledge of the system dynamics is required. It is well known that modeling and identification for an unknown nonlinear system are most often time consuming, which require iterative execution of model assumption, parameter identification and model validation. To obviate the dependence on complete knowledge of the system dynamics, a direct adaptive optimal control algorithm was presented, in which a discretized version Bellman equation (BE) was introduced to approximate the solution to the HJB equation without using the information of the drift dynamics of system [17]. This method was extended to optimal tracking control of nonlinear constrained-input systems [18], and named integral reinforcement learning (IRL). A simultaneous policy update algorithm (SPUA) was introduced for solving $H_\infty$ control problem [19], it was a generalization of the IRL algorithm in $H_\infty$ control. SUPA is an on-policy learning method, which have several drawbacks for solving the $H_\infty$ control problem, such as disturbance signal should be adjustable and exploration space is limited. To overcome the drawbacks, an off-policy policy iteration method was presented by using arbitrarily behavior policy in data sampling on the state and input domain for optimal control design [20]-[22]. Subsequently, off-policy learning methods for $H_\infty$ control problems were proposed for partially unknown nonlinear systems and completely unknown nonlinear systems, respectively [23]-[25]. The off-policy learning method was employed to solve $H_\infty$ tracking control [6]. It is of interest to note that the on-policy and off-policy PI algorithms mentioned above are in essence the Newton method, Newton's iterative approach is locally optimized, and finding suitable initial parameters or developing global methods is a difficult problem. In recent years, some new methods, such as integration scheme of policy iteration and value iteration, model-free $\lambda$-PI algorithm, and costate-supplement ADP algorithm, have been proposed to improve convergence [26]-[28].

Till present, the development of policy iteration algorithms and theories for $H_\infty$ tracking control design of continuous-time nonlinear system is still an open issue, which promotes this research. It is worth noting that damped Newton method, which leverages a non-unit step-size in the Newton direction by multiplying a damped parameter $\delta$ is an effective algorithm for solving equations. Although damped Newton's method makes the convergence slower than Newton's method, damped Newton method could be globally convergent. Therefore, the damped Newton method has the advantage of enhancing the robustness of the convergence of solving process with regarding to the initial guess [29].

In this paper, the $H_\infty$ tracking control design of nonlinear continuous-time systems is considered, and a novel policy iteration algorithm based on damped Newton method is developed for solving tracking HJI equation, named $\delta$-policy iteration ($\delta$-PI) algorithm. A generalized tracking Bellman equation is derived by constructing a damped Newton iteration operator equation. The generalized tracking Bellman equation is an extension of tracking Bellman equation introduced in [6]. And then, by iterating on the generalized tracking Bellman equation, one can obtain the approximate optimal solution of the tracking HJI equation. On-policy and off-policy $\delta$-PI reinforcement learning methods are given respectively.

The organization of this paper is as follows. The continuous-time nonlinear $H_\infty$ tracking control problem formulation and preliminaries are presented in section II. The Newton method based on-policy PI algorithm is presented in section III. The damped Newton method based on-policy and off-policy $\delta$-PI RL methods, and neural-network based implementation schemes are provided in section IV. The effectiveness of the $\delta$-PI RL algorithm is illustrated with a simulation example in section V. Finally, section VI concludes the paper.

## II. PROBLEM FORMULATION AND PRELIMINARIES

In this section, the background knowledge reviews and preliminaries of $H_\infty$ tracking control problem are provided. Consider a class of affine in control input nonlinear continuous-time system defined as follows

$$\dot{x} = f(x) + g(x)u + k(x)d \quad (1)$$

where $x \in \Omega \subset \Re^n$ is state vector of the system, and $d \in \Re^q$ and $u \in \Re^m$ represent disturbance input vector and control input vector respectively. $f(x) \in \Re^n$ is the internal dynamics, $g(x) \in \Re^{n \times m}$ and $k(x) \in \Re^{n \times q}$ are the input-to-state matrix and the disturbance coefficient matrix, respectively. On a compact set $\Omega$, $f(x)$, $g(x)$ and $k(x)$ meet locally Lipschitz continuity and $f(0) = 0$, and that the system (1) is stabilizable.

$H_\infty$ tracking control is a robust tracking control method, whose goal is to reduce the impact of external disturbance when system (1) tracks a reference trajectory $r(t) \subset \Re^n$.

Define the tracking error $e_d$ as follows.

$$e_d \triangleq x(t) - r(t) \quad (2)$$

*Assumption 1*: Reference trajectory $r(t)$ is bounded and there exists a Lipschitz continuous command generator function $h_d(r(t)) \in \Re^n$ and $h_d(0) = 0$, such that

$$\dot{r}(t) = h_d(r(t)) \quad (3)$$

By using equations (1)-(3), the error dynamics of the tracking system (1) can be obtained as

$$\dot{e}_d = f(x) + g(x)u + k(x)d - h_d(r) \quad (4)$$

For any $d \in L_2[0, \infty)$, a more general version of disturbance attenuation condition of $H_\infty$ tracking problem is given as follows [6].

$$\frac{\int_t^\infty e^{-\alpha(\tau-t)}(e_d^T Q e_d + u^T R u) d\tau}{\int_t^\infty e^{-\alpha(\tau-t)}(d^T d) d\tau} \leq \gamma^2 \quad (5)$$

where $\alpha > 0$ is a discount factor, $Q > 0$ and $R > 0$ are positive definite symmetric matrix and positive definite diagonal matrix respectively, and $\gamma > 0$ is a predetermined disturbance attenuation level.

The $H_\infty$ optimal control is interested in finding the smallest $\gamma^*$, such that there exists a solution that satisfies the disturbance attenuation condition (5). However, it is impossible to get the smallest value of $\gamma^*$ for general nonlinear systems, thus it is assumed that $\gamma > \gamma^*$ in this paper [30].

*Definition 1* (Bounded $L_2$-Gain): If system (1) satisfies the disturbance attenuation condition (5), then the $L_2$-gain of system (1) is less than or equal to $\gamma$.

*Remark 1*: The use of discount factor of exponential decay type is crucial in (5). This is because the control input in tracking problems usually does not converge to zero, therefore punishing the energies of the tracking error and the whole control input without a discount factor of exponential decay type will make the performance index function unbounded.

The purpose of the tracking problem $H_\infty$ control is to find a control strategy $u = u(e_d, r)$ that depends on the tracking error $e_d$ and the reference trajectory $r(t)$, such that the closed-loop error dynamics (4) satisfies the disturbance attenuation condition (5), and the tracking error meets local asymptotic stability when $d = 0$.

Define the extended system state variable $X(t) = [e_d(t)^T \ r(t)^T]^T \in \Omega \subset \mathbb{R}^{2n}$, then the augmented system which is consisted of tracking error dynamics and reference signal command generator function is given as follows.

$$\dot{X}(t) = F(X(t)) + G(X(t))u(t) + K(X(t))d(t) \quad (6)$$

where $F(X(t)) = \begin{bmatrix} f(e_d + r) - h_d(r) \\ h_d(r) \end{bmatrix}$, $G(X) = \begin{bmatrix} g(e_d + r) \\ 0 \end{bmatrix}$,

and $K(X(t)) = \begin{bmatrix} k(e_d + r) \\ 0 \end{bmatrix}$. Based upon the extended system (6), the attenuation condition of disturbance (5) becomes

$$\int_t^\infty e^{-\alpha(\tau-t)}(X^T Q_T X + u^T R u) d\tau \leq \gamma^2 \int_t^\infty e^{-\alpha(\tau-t)}(d^T d) d\tau \quad (7)$$

where $Q_T = \begin{bmatrix} Q & 0 \\ 0 & 0 \end{bmatrix}$.

Define the discount performance index function of the tracking problem as

$$J(u, d) = \int_t^\infty e^{-\alpha(\tau-t)}(X^T Q_T X + u^T R u - \gamma^2 d^T d) d\tau \quad (8)$$

Note that for the $H_\infty$ tracking problem, solving the control strategy which satisfies the bounded $L_2$-gain condition is equivalent to solving a differential game problem of the augmented system, i.e., a min-max optimization problem. As is well known, the two-person zero-sum differential games problem is closely related to $H_\infty$ control [31], that is, the solvability of the $H_\infty$ control problem is equivalent to the solvability of the following two-person zero-sum differential games problem.

$$V^*(X) = J(u^*, d^*) = \min_u \max_d J(u, d) \quad (9)$$

where $V^*(X)$ is the optimal value function. If the Nash condition given below holds, then the two-person zero-sum differential games problem has a unique saddle point solution.

$$V^*(X) = \min_u \max_d J(u, d) = \max_d \min_u J(u, d) \quad (10)$$

Calculate the differential on both sides of (8), and note that $V(X) = J(u, d)$, then the tracking Bellman equation is obtained as follows.

$$H(V, u, d) \triangleq$$
$$X^T Q_T X + u^T R u - \gamma^2 d^T d - \alpha V + \left(\frac{\partial V}{\partial X}\right)^T (F + Gu + Kd) = 0 \quad (11)$$

Employing the first-order necessary conditions $\partial H(V^*, u, d)/\partial d = 0$ and $\partial H(V^*, u, d)/\partial u = 0$ for the min-max optimization problem, one can obtain the optimal control input and disturbance input as

$$u^* = -\frac{1}{2} R^{-1} G^T V_X^* \quad (12)$$

$$d^* = \frac{1}{2\gamma^2} K^T V_X^* \quad (13)$$

where $V_X^* = \nabla V^* = \partial V^*/\partial X$, $V^*$ is the optimal value function described in (9). By substituting the optimal control policy (12) and the optimal disturbance signal (13) into (11), one can obtain the tracking problem HJI equation, as follows.

$$H(V^*, u^*, d^*) \triangleq X^T Q_T X - \alpha V^* + (V_X^*)^T F$$
$$-\frac{1}{4}(V_X^*)^T G R^{-1} G^T V_X^* + \frac{1}{4\gamma^2}(V_X^*)^T K K^T V_X^* = 0 \quad (14)$$

*Lemma 1 ([6, Theorem 1]):* Take the $H_\infty$ tracking control problem as a two-person zero-sum differential games problem of the augmented system with performance index function (8), then the control strategy pairs defined in (12) and (13) offer as Nash equilibrium saddle-point solutions.

*Lemma 2 ([6, Theorem 2]):* Assume that the tracking HJI equation (14) has a continuous positive-semidefinite solution $V^*$, then the control input given in (12) makes the $L_2$-gain of the closed-loop augmented system (6) less than or equal to $\gamma$.

Reference [6] proved the existence of the upper bound $\alpha^*$ on the discount factor $\alpha$ to ensure the stability of the solution of the tracking HJI equation (14). In practice, a large weighting design matrix $Q$ and/or a very small discount factor $\alpha$ can always be chose to ensure that $\alpha \leq \alpha^*$.

## III. NEWTON'S ITERATIVE METHOD

Note that the tracking Bellman equation (11) is linear with respect to the value function $V$, whereas the tracking HJI equation (14) is nonlinear with respect to the optimal value function $V^*$. Therefore, solving the value function $V$ in the tracking Bellman equation is easier than solving $V^*$ in the tracking HJI equation. The PI algorithm for tracking problems proposed in [6] decomposes the solution of the tracking HJI equation into a series of solutions of tracking Bellman equation, instead of directly solving from the tracking HJI equation. The PI algorithm is as follows.

---

**Algorithm 1** PI for $H_\infty$ Tracking Control.

Step 1: Set $i = 0$, give an initial admissible control $u_i$ and a disturbance signal $d_i$.

Step 2: Solve $V_{i+1}(X)$ with $V_{i+1}(0) = 0$ using the tracking Bellman equation below

$$X^T Q_T X + u_i^T R u_i - \gamma^2 d_i^T d_i - \alpha V_{i+1} + \left(\frac{\partial V_{i+1}}{\partial X}\right)^T (F + G u_i + K d_i) = 0 \quad (15)$$

Step 3: Update the disturbance policy and control policy using

$$u_{i+1} = -\frac{1}{2} R^{-1} G^T \frac{\partial V_{i+1}}{\partial X} \quad (16)$$

$$d_{i+1} = \frac{1}{2\gamma^2} K^T \frac{\partial V_{i+1}}{\partial X} \quad (17)$$

Step 4: Set $i = i + 1$, move to Step 2 and go on iteration, until $\|V_{i+1} - V_i\|_\Omega \leq \varepsilon$, $\varepsilon$ is a small positive number.

In fact, the convergence of Algorithm 1 can be demonstrated by proving that the iterations of (15)-(17) are essentially Newton iteration method that converge to the unique solution of the tracking HJI equation (14).

Firstly, consider a Banach space $\mathbb{V} = \{V(X)|V(X): \Omega \to \Re\}$, and define the following mapping $\mathcal{G}: \mathbb{V} \to \mathbb{V}$ on $\mathbb{V}$:

$$\mathcal{G}(V) \triangleq X^T Q_T X - \alpha V + V_X^T F - \frac{1}{4} V_X^T G R^{-1} G^T V_X + \frac{1}{4\gamma^2} V_X^T K K^T V_X \quad (18)$$

The Fréchet derivative of mapping $\mathcal{G}(V)$ is denoted as $\mathcal{G}'(V)$, and the Fréchet differential is written as $\mathcal{G}'(V)W$, where $W \in \widetilde{\mathbb{V}} \subset \mathbb{V}$ and $\widetilde{\mathbb{V}}$ is a neighborhood of $V$. Since the Fréchet derivative is usually difficult to calculate directly, the Gâteaux derivative is an alternative. The Gâteaux differential $\mathcal{L}(W)$ of $\mathcal{G}$ at the point $V$ is [32]

$$\mathcal{L}(W) = \lim_{s \to 0} \frac{\mathcal{G}(V + sW) - \mathcal{G}(V)}{s} \quad (19)$$

where $\mathcal{L}$ is the Gâteaux derivative of $\mathcal{G}$ at the point $V$, $W \in \widetilde{\mathbb{V}}$, $\|W\| = 1$, $s$ is a real number. Equation (19) gives a method to calculate Gâteaux derivative, rather than Fréchet derivative. The *Lemma 3* gives the relationship between Gâteaux derivative and Fréchet derivative, and *Lemma 4* provides the method of calculation for Fréchet differential by using Gâteaux differential.

*Lemma 3 [33]:* If Gâteaux derivative $\mathcal{L}$ exists in some neighborhood of $V$, and if $\mathcal{L}$ is continuous at the point $V$, then $\mathcal{L}$ is also a Fréchet derivative at the point $V$.

*Lemma 4:* On the Banach space $\mathbb{V}$, $\forall V \in \mathbb{V}$, consider the mapping $\mathcal{G}$ defined in (18), the Fréchet differential of $\mathcal{G}$ can be computed as

$$\mathcal{G}'(V)W = \nabla(W)^T F - \alpha(W) - \frac{1}{4} \nabla(V)^T G R^{-1} G^T \nabla(W) - \frac{1}{4} \nabla(W)^T G R^{-1} G^T \nabla(V) + \frac{1}{4\gamma^2} \nabla(V)^T K K^T \nabla(W) + \frac{1}{4\gamma^2} \nabla(W)^T K K^T \nabla(V) \quad (20)$$

where $\nabla(\cdot) = \partial(\cdot)/\partial X$.

*Proof:* The molecule of (19) can be calculated as
$$\mathcal{G}(V + sW) - \mathcal{G}(V) = X^T Q X + \nabla(V + sW)^T F$$
$$-\alpha(V + sW) - \frac{1}{4} \nabla(V + sW)^T G R^{-1} G^T \nabla(V + sW)$$
$$+ \frac{1}{4\gamma^2} \nabla(V + sW)^T K K^T \nabla(V + sW) - \left[X^T Q X + \nabla(V)^T F - \alpha(V) - \frac{1}{4} \nabla(V)^T G R^{-1} G^T \nabla(V) + \frac{1}{4\gamma^2} \nabla(V)^T K K^T \nabla(V)\right] =$$
$$s \nabla(W)^T F - \alpha s(W) - \frac{1}{4} s \nabla(V)^T G R^{-1} G^T \nabla(W)$$
$$- \frac{1}{4} s \nabla(W)^T G R^{-1} G^T \nabla(V) - \frac{1}{4} s^2 \nabla(W)^T G R^{-1} G^T \nabla(W)$$
$$+ \frac{1}{4\gamma^2} s \nabla(V)^T K K^T \nabla(W) + \frac{1}{4\gamma^2} s \nabla(W)^T K K^T \nabla(V)$$
$$+ \frac{1}{4\gamma^2} s^2 \nabla(W)^T K K^T \nabla(W) \quad (21)$$

Then
$$\mathcal{L}(W) = \lim_{s \to 0} \frac{\mathcal{G}(V + sW) - \mathcal{G}(V)}{s}$$
$$= \nabla(W)^T F - \alpha(W) - \frac{1}{4} \nabla(V)^T G R^{-1} G^T \nabla(W)$$
$$- \frac{1}{4} \nabla(W)^T G R^{-1} G^T \nabla(V) + \frac{1}{4\gamma^2} \nabla(V)^T K K^T \nabla(W)$$
$$+ \frac{1}{4\gamma^2} \nabla(W)^T K K^T \nabla(V) \quad (22)$$

On the other hand, use the same method as in [15], it can be proven that $\mathcal{L}$ is a continuous mapping. According to *Lemma 3*, $\mathcal{L}(W)$ is the Fréchet differential, $\mathcal{L}$ is the Fréchet derivative at $V$. This completes the proof. □

The following theorem will prove that the Algorithm 1 given above is mathematically equivalent to Newton's iteration in the Banach space $\mathbb{V}$.

*Theorem 1*: Define mapping $\mathcal{T}: \mathbb{V} \to \mathbb{V}$ on Banach space
$$\mathcal{T}(V) = V - \left(\mathcal{G}'(V)\right)^{-1} \mathcal{G}(V) \quad (23)$$

Then, the following Newton's iteration equation with (16) and (17) is equivalent to the PI algorithm from (15) to (17).
$$V_{i+1} = \mathcal{T}(V_i) = V_i - \left(\mathcal{G}'(V_i)\right)^{-1} \mathcal{G}(V_i) \quad (24)$$

*Proof:* Equation (24) can be rewritten as
$$\mathcal{G}'(V_i) V_{i+1} = \mathcal{G}'(V_i) V_i - \mathcal{G}(V_i) \quad (25)$$

By using *Lemma 4*, one can get
$$\mathcal{G}'(V_i) V_{i+1} = \nabla(V_{i+1})^T F - \alpha(V_{i+1})$$
$$- \frac{1}{4} \nabla(V_i)^T G R^{-1} G^T \nabla(V_{i+1}) - \frac{1}{4} \nabla(V_{i+1})^T G R^{-1} G^T \nabla(V_i)$$
$$+ \frac{1}{4\gamma^2} \nabla(V_i)^T K K^T \nabla(V_{i+1}) + \frac{1}{4\gamma^2} \nabla(V_{i+1})^T K K^T \nabla(V_i) \quad (26)$$

$$\mathcal{G}'(V_i) V_i = \nabla(V_i)^T F - \alpha(V_i) - \frac{1}{4} \nabla(V_i)^T G R^{-1} G^T \nabla(V_i)$$
$$- \frac{1}{4} \nabla(V_i)^T G R^{-1} G^T \nabla(V_i) + \frac{1}{4\gamma^2} \nabla(V_i)^T K K^T \nabla(V_i)$$
$$+ \frac{1}{4\gamma^2} \nabla(V_i)^T K K^T \nabla(V_i) \quad (27)$$

From (18), (16), and (17), one has
$$\mathcal{G}(V_i) = X^T Q_T X + \nabla(V_i)^T F - \alpha(V_i) - \frac{1}{4} \nabla(V_i)^T G R^{-1} G^T \nabla(V_i)$$
$$+ \frac{1}{4\gamma^2} \nabla(V_i)^T K K^T \nabla(V_i) \quad (28)$$

$$u_i = -\frac{1}{2} R^{-1} G^T \nabla V_i \quad (29)$$

$$d_i = \frac{1}{2\gamma^2} K^T \nabla V_i \quad (30)$$

Substituting (26)-(30) into (25), yields the tracking Bellman (15)
$$-\alpha(V_{i+1}) + \nabla(V_{i+1})^T [F + G u_i + K d_i] =$$
$$-X^T Q_T X - u_i^T R u_i + \gamma^2 d_i^T d_i \quad (31)$$

The Newton iterative equation (24) with (16) and (17) is equivalent to the Algorithm 1 from (15) to (17). This completes the proof. □

The convergence of sequence $\{V_i\}$ given in (24) to the unique solution of the fixed-point equation $V^* = \mathcal{T}(V^*)$ can be guaranteed by Kantorovich's theorem [34], then the solution of $\mathcal{G}(V^*) = 0$ is obtained, i.e., the solution of the tracking HJI equation (14).

## IV. $\delta$-PI Algorithm for Solving the Tracking HJI Equation

The damped Newton method is an effective method for solving nonlinear equations, which can adjust the calculation step in the Newton direction through damped parameter $\delta$. The main advantage of the damped Newton method is that it improves the robustness of the solution procedure with respect to the initial conditions [29]. In this section, a damped Newton iteration based RL method is proposed for $H_\infty$ tracking problems, named $\delta$-PI algorithm, which can be used to pursue the solutions of tracking HJI equation and obtain the $H_\infty$ tracking controller. On-policy learning method and off-policy learning method are presented, respectively. Then, a NN-based implementation scheme is employed to implement the off-policy $\delta$-PI algorithm without making use of any information of the system dynamics.

### A. On-policy $\delta$-PI Algorithm

Based on the damped Newton method, the on-policy $\delta$-PI is given in Algorithm 2.

**Algorithm 2** On-policy $\delta$-PI for $H_\infty$ Tracking Control.

Step 1: Set $i = 0$, give an initial cost function $V_0$, an initial admissible control $u_i$ and a disturbance signal $d_i$.

Step 2: Solve $V_{i+1}(X)$ with $V_{i+1}(0) = 0$ from the generalized tracking Bellman equation (32).
$$-\alpha(V_{i+1}) + \nabla(V_{i+1})^T[F + Gu_i + Kd_i] = \\ (1-\delta)\{\nabla(V_i)^T[F + Gu_i + Kd_i] - \alpha V_i\} \\ -\delta[X^T Q_T X + u_i^T R u_i - \gamma^2 d_i^T d_i] \quad (32)$$

Step 3: Update the disturbance policy and control policy using
$$u_{i+1} = -\frac{1}{2}R^{-1}G^T \frac{\partial V_{i+1}}{\partial X} \quad (33)$$

$$d_{i+1} = \frac{1}{2\gamma^2}K^T \frac{\partial V_{i+1}}{\partial X} \quad (34)$$

Step 4: Set $i = i + 1$, move to Step 2 and go on iteration, until a certain stopping criterion is met.

Define a new mapping $\mathcal{T}_\delta: \mathbb{V} \to \mathbb{V}$ on Banach space $\mathbb{V}$
$$\mathcal{T}_\delta(V) = V - \delta\big(\mathcal{G}'(V)\big)^{-1}\mathcal{G}(V) \quad (35)$$
where $0 < \delta \le 1$.

*Theorem 2*: Given the damped Newton mapping $\mathcal{T}_\delta$ shown in (35), then the following damped Newton iteration with (33) and (34) is equivalent to the on-policy $\delta$-PI algorithm from (32) to (34).
$$V_{i+1} = \mathcal{T}_\delta(V_i) \quad (36)$$

*Proof*: Substituting (35) into (36) and making some changes, one gets

$$\mathcal{G}'(V_i)V_{i+1} = \mathcal{G}'(V_i)V_i - \delta\mathcal{G}(V_i). \quad (37)$$

Substituting (26), (27), and (18) into (37) yields
$$(V_{i+1})^T F - \alpha(V_{i+1}) - \frac{1}{4}\nabla(V_i)^T G R^{-1} G^T \nabla(V_{i+1}) \\ -\frac{1}{4}\nabla(V_{i+1})^T G R^{-1} G^T \nabla(V_i) + \frac{1}{4\gamma^2}\nabla(V_i)^T K K^T \nabla(V_{i+1}) \\ + \frac{1}{4\gamma^2}\nabla(V_{i+1})^T K K^T \nabla(V_i) = \nabla(V_i)^T F - \alpha(V_i) \\ -\frac{1}{4}\nabla(V_i)^T G R^{-1} G^T \nabla(V_i) - \frac{1}{4}\nabla(V_i)^T G R^{-1} G^T \nabla(V_i) \\ + \frac{1}{4\gamma^2}\nabla(V_i)^T K K^T \nabla(V_i) + \frac{1}{4\gamma^2}\nabla(V_i)^T K K^T \nabla(V_i) \\ -\delta\left[X^T Q_T X + \nabla(V_i)^T F - \alpha(V_i) - \frac{1}{4}\nabla(V_i)^T G R^{-1} G^T \nabla(V_i) + \frac{1}{4\gamma^2}\nabla(V_i)^T K K^T \nabla(V_i)\right] \quad (38)$$

Make some mathematical manipulate, one can obtain
$$-\alpha(V_{i+1}) + \nabla(V_{i+1})^T[F + Gu_i + Kd_i] = -(1-\delta)\alpha V_i \\ +(1-\delta)\nabla(V_i)^T F + (1-\delta)\left[-\frac{1}{2}\nabla(V_i)^T G R^{-1} G^T \nabla(V_i)\right] \\ +(1-\delta)\left[\frac{1}{2\gamma^2}\nabla(V_i)^T K K^T \nabla(V_i)\right] - \delta\left[X^T Q_T X + \frac{1}{4}\nabla(V_i)^T G R^{-1} G^T \nabla(V_i) - \frac{1}{4\gamma^2}\nabla(V_i)^T K K^T \nabla(V_i)\right] \quad (39)$$

Substitute (33) and (34) into (39), one can obtain the generalized tracking Bellman equation (32). This means that the damped Newton iteration given by (36) with (33) and (34) is equivalent to the on-policy $\delta$-PI algorithm from (32) to (34). This completes the proof. □

*Remark 2*: Compared with tracking Bellman equation (11), the reinforcement term of the generalized tracking Bellman equation (32) is multiplied by the damped coefficient $\delta$. When $\delta = 1$, (32) degenerates to tracking Bellman equation (15), and on-Policy $\delta$-PI algorithm becomes PI algorithm as shown in Algorithm 1.

*Remark 3:* The generalized tracking Bellman equation (32) is linear partial differential equation just like the tracking bellman equation (11).

The complete knowledge of the augmented system dynamics is required for solving $V_{i+1}(X)$ from (32). Multiplying the two sides of (32) by $e^{-\alpha(\tau-t)}$ and integrating in the time interval $(t, t+T]$, one can obtain a discrete version of the generalized tracking Bellman equation, as follows.
$$e^{-\alpha T}V_{i+1}\big(X(t+T)\big) - V_{i+1}\big(X(t)\big) = \\ -\delta \int_t^{t+T} e^{-\alpha(\tau-t)}([X^T Q_T X + u_i^T R u_i - \gamma^2 d_i^T d_i])d\tau \\ +(1-\delta)\big[e^{-\alpha T}V_i\big(X(t+T)\big) - V_i\big(X(t)\big)\big] \quad (40)$$

Equations (40) and (32) have the same solution while $V_{i+1}(0) = 0$, which can be proved in line with [15]. Using (40) instead of (32) in Algorithm 2, one can obtain a partially model-free $\delta$-PI algorithm. The NN implementation scheme of Algorithm 2 can be referred to [15].

### B. Off-policy $\delta$-PI Algorithm

Algorithm 2 is an on-policy learning method, in which the disturbance signals need to be adjustable when implemented online, and it is necessary to know the dynamic information of the system. This is not in line with the reality. To overcome this drawback, motivated by [20]-[25], an off-policy $\delta$-PI algorithm based on the generalized tracking Bellman equation is proposed to solve the tracking HJI equation through learning arbitrary behavior policies, and none of the prior information of the dynamics of the system is required. Then, three NNs are used to achieve the off-policy $\delta$-PI tracking algorithm proposed.

Firstly, the augmented system dynamics (6) can be rewritten as

$$\dot{X} = F + Gu_i + Kd_i + G(u - u_i) + K(d - d_i) \quad (41)$$

where $u \in \Re^m$ and $d \in \Re^q$ are behavior policy and actual disturbance, respectively. $u_i \in \Re^m$ and $d_i \in \Re^q$ are policies to be evaluated and updated. Let $V_{i+1}(x)$ be the solution of generalized tracking Bellman equation (32), and differentiating $V_{i+1}(x)$ along with the augmented system dynamics (41) yields

$$\dot{V}_{i+1} = (\nabla V_{i+1})^T(F + Gu_i + Kd_i) + (\nabla V_{i+1})^T G(u - u_i)$$
$$+ (\nabla V_{i+1})^T K(d - d_i) \quad (42)$$

Substituting (32) into the above equation yields

$$\dot{V}_{i+1} = -\delta[X^T Q_T X + u_i^T R u_i - \gamma^2 d_i^T d_i]$$
$$+ (1-\delta)\{\nabla(V_i)^T[F + Gu_i + Kd_i] - \alpha V_i\} + \alpha(V_{i+1})$$
$$+ (\nabla V_{i+1})^T G(u - u_i) + (\nabla V_{i+1})^T K(d - d_i) \quad (43)$$

Substituting the augmented system dynamics (41) into (43) yields

$$\dot{V}_{i+1} = -\delta[X^T Q_T X + u_i^T R u_i - \gamma^2 d_i^T d_i]$$
$$+ (1-\delta)\nabla(V_i)^T[\dot{X} - G(u - u_i) - K(d - d_i)]$$
$$-(1-\delta)\alpha V_i + \alpha(V_{i+1}) + (\nabla V_{i+1})^T G(u - u_i)$$
$$+ (\nabla V_{i+1})^T K(d - d_i)$$
$$= -\delta[X^T Q_T X + u_i^T R u_i - \gamma^2 d_i^T d_i]$$
$$+ (1-\delta)\nabla(V_i)^T \dot{X} - (1-\delta)\nabla(V_i)^T G(u - u_i)$$
$$-(1-\delta)\nabla(V_i)^T K(d - d_i) - (1-\delta)\alpha V_i + \alpha(V_{i+1})$$
$$+ (\nabla V_{i+1})^T G(u - u_i) + (\nabla V_{i+1})^T K(d - d_i) \quad (44)$$

Substituting $d_i = \frac{1}{2\gamma^2} K^T \nabla V_i$, $u_i = -\frac{1}{2} R^{-1} G^T \nabla V_i$, $d_{i+1} = \frac{1}{2\gamma^2} K^T \nabla V_{i+1}$, and $u_{i+1} = -\frac{1}{2} R^{-1} G^T \nabla V_{i+1}$ into (44) yields an off-policy generalized tracking Bellman equation.

$$\dot{V}_{i+1} - \alpha(V_{i+1}) = -\delta[X^T Q_T X + u_i^T R u_i - \gamma^2 d_i^T d_i]$$
$$+ (1-\delta)[\nabla(V_i)^T \dot{X} - \alpha V_i] + (1-\delta)2u_i^T R(u - u_i)$$
$$-(1-\delta)2\gamma^2 d_i^T(d - d_i) - 2u_{i+1}^T R(u - u_i)$$
$$+ 2\gamma^2 d_{i+1}^T(d - d_i) \quad (45)$$

From the derivation process of (45), it can be seen that (45) and the generalized tracking Bellman equation (32) have the same solution. For space reasons the proof is omitted in this paper. Multiplying the two sides of (45) by $e^{-\alpha(\tau-t)}$ and integrating in the time interval $(t, t+T]$, one can obtain a discrete version of the off-policy generalized tracking Bellman equation, as follows.

$$e^{-\alpha T} V_{i+1}(X(t+T)) - V_{i+1}(X(t)) =$$
$$-\delta \int_t^{t+T} e^{-\alpha(\tau-t)}([X^T Q_T X + u_i^T R u_i - \gamma^2 d_i^T d_i])d\tau$$
$$+ (1-\delta)[e^{-\alpha T} V_i(X(t+T)) - V_i(X(t))]$$
$$+ (1-\delta) \int_t^{t+T} e^{-\alpha(\tau-t)} 2u_i^T R(u - u_i)d\tau$$
$$-(1-\delta) \int_t^{t+T} e^{-\alpha(\tau-t)} 2\gamma^2 d_i^T(d - d_i)d\tau$$
$$- \int_t^{t+T} e^{-\alpha(\tau-t)} 2u_{i+1}^T R(u - u_i)d\tau$$
$$+ \int_t^{t+T} e^{-\alpha(\tau-t)} 2\gamma^2 d_{i+1}^T(d - d_i)d\tau \quad (46)$$

It is observed from (46) that arbitrary input signals $u$ and $d$ can be used for learning the value function $V_{i+1}$, rather than the policies $u_i$ and $d_i$ to be evaluated. Then, replacing (32) in Algorithm 2 with (46), one can obtains the off-policy $\delta$-PI Algorithm, as shown in Algorithm 3.

---

**Algorithm 3** Off-Policy $\delta$-PI for $H_\infty$ Tracking Control Problem.

---

Step 1: Use the behavior policy $u$ and the actual disturbance $d$ to collect $N$ system data which contain system state, disturbance input and control input at different sampling time interval.

Step 2: Set $i = 0$, give an initial cost function $V_0$ with $V_0(0) = 0$, and initial control and disturbance policies $u_0$ and $d_0$.

Step 3: Reuse the collected data to solve (46) for $V_{i+1}(x)$, $u_{i+1}$ and $d_{i+1}$, with $V_{i+1}(0) = 0$.

Step 4: If a stop iteration condition is met, stop iteration and output $V_{i+1}(x)$ as the approximate optimal solution of tracking HJI equation (14), output $u_{i+1}$ as the approximate optimal control input, i.e., $V^*(x) = V_{i+1}(x)$ and $u^* = u_{i+1}$, else set $i = i + 1$, move to Step 3 and go on iteration.

---

Algorithm 3 consists of two independent stages. In the first stage, behavior strategies are used to collect system data. In the second stage, the data collected in the first stage are reused for policies and value function update. The implementation of the second stage update can be carried out without using any knowledge of the system dynamics, that is, Algorithm 3 is a model-free algorithm. When $\delta = 1$, (46) degenerates to the off-policy RL Bellman equation as shown in [6].

*C. NN-Based Implement of Algorithm 3*

Equation (46) is a scalar equation that can be solved in the sense of least-squares after sampling a sufficient number of data from the augmented system. Next, a NN-based actor-critic structure is introduced to implement Algorithm 3, in which two actor NNs and one critic NN are adopted to learn the value function, disturbance policy and control policy, respectively. The three NNs are given as follows

$$\hat{V}_{i+1}(X) = W_c^T \rho(X) \quad (47)$$
$$\hat{u}_{i+1}(X) = W_a^T \phi(X) \quad (48)$$
$$\hat{d}_{i+1}(X) = W_d^T \varphi(X) \quad (49)$$

where $\rho(X) = [\rho_1(X), \ldots, \rho_{L_1}(X)]^T \in \Re^{L_1}$, $\phi(X) = [\phi_1(X), \ldots, \phi_{L_2}(X)]^T \in \Re^{L_2}$, and $\varphi(X) = [\varphi_1(X), \ldots, \varphi_{L_3}(X)]^T \in \Re^{L_3}$ are the linearly independent basis functions of the three NNs, respectively, which are all defined on $\Omega \subset \Re^n$. $W_c \in \Re^{L_1}$, $W_a \in \Re^{L_2 \times m}$, and $W_d \in \Re^{L_3 \times q}$. Based on high-order Weierstrass approximation theories [35], there exists a complete basis set, that the approximation error of NNs tends to zero uniformly when $L_1, L_2,$ and $L_3 \to \infty$.

Substituting (47)-(49) into (46) yields

$$e^{-\alpha T} W_c^T \rho(X(t+T)) - W_c^T \rho(X(t))$$
$$+2\sum_{j=1}^m r_j \int_t^{t+T} e^{-\alpha(\tau-t)} W_{a,j}^T \phi(X) v_j^1 \, d\tau$$
$$-2\gamma^2 \sum_{k=1}^q \int_t^{t+T} e^{-\alpha(\tau-t)} W_{d,k}^T \varphi(X) v_k^2 \, d\tau =$$
$$-\delta \int_t^{t+T} e^{-\alpha(\tau-t)} ([X^T Q_T X + u_i^T R u_i - \gamma^2 d_i^T d_i]) d\tau$$
$$+(1-\delta)[e^{-\alpha T} V_i(X(t+T)) - V_i(X(t))]$$
$$+(1-\delta) \int_t^{t+T} e^{-\alpha(\tau-t)} 2 u_i^T R(u - u_i) d\tau$$
$$-(1-\delta) \int_t^{t+T} e^{-\alpha(\tau-t)} 2\gamma^2 d_i^T (d - d_i) d\tau \quad (50)$$

where $v^1 = [v_1^1, \ldots, v_m^1]^T = u - u_i$ and $v^2 = [v_1^2, \ldots, v_q^2]^T = d - d_i$. $W_{a,j}$ is the $j^{th}$ column of matrix $W_a$, $W_{d,k}$ is the $k^{th}$ column of matrix $W_d$.

Note that (50) is linear in the NNs weight vectors $W_c$, $W_a$ and $W_d$. Define

$$W = [W_c^T, W_{a,1}^T, \ldots, W_{a,m}^T, W_{d,1}^T, \ldots, W_{d,q}^T]^T \quad (51)$$

$$\omega(t) = \begin{bmatrix} e^{-\alpha T} \rho(X(t+\Delta t)) - \rho(X(t)) \\ 2r_1 \int_t^{t+T} e^{-\alpha(\tau-t)} \phi(X) v_1^1 \, d\tau \\ \vdots \\ 2r_m \int_t^{t+T} e^{-\alpha(\tau-t)} \phi(X) v_m^1 \, d\tau \\ -2\gamma^2 \int_t^{t+T} e^{-\alpha(\tau-t)} \varphi(X) v_1^2 \, d\tau \\ \vdots \\ -2\gamma^2 \int_t^{t+T} e^{-\alpha(\tau-t)} \varphi(X) v_q^2 \, d\tau \end{bmatrix} \quad (52)$$

$$\lambda(t) = -\delta \int_t^{t+T} e^{-\alpha(\tau-t)} ([X^T Q_T X + u_i^T R u_i - \gamma^2 d_i^T d_i]) d\tau$$
$$+(1-\delta)[e^{-\alpha T} V_i(X(t+T)) - V_i(X(t))]$$
$$+(1-\delta) \int_t^{t+T} e^{-\alpha(\tau-t)} 2 u_i^T R(u - u_i) d\tau$$
$$-(1-\delta) \int_t^{t+T} e^{-\alpha(\tau-t)} 2\gamma^2 d_i^T (d - d_i) d\tau \quad (53)$$

Then (50) can be rewritten as
$$\lambda(t) = W^T \omega(t) \quad (54)$$

Note that (54) is linear with respect to the weight $W$ of the three NNs, thus $W$ can be solved in the sense of least-squares. Because of $W \in \mathbb{R}^{L_1 + m \times L_2 + q \times L_3}$, therefore one needs to collect $N > L_1 + m \times L_2 + q \times L_3$ augmented system data about system state, disturbance and control input from $t_1$ to $t_N$ in the state space. Then, for the given value function $V_i$, evaluated policies $u_i$ and $d_i$, using this information to calculate (52) and (53) at the $N$ different sampling data point, one can get

$$\Omega = [\omega(t_1), \ldots, \omega(t_N)] \quad (55)$$
$$\Lambda = [\lambda(t_1), \ldots, \lambda(t_N)]^T \quad (56)$$

The least-squares solution of the NNs weight $W$ is
$$W = (\Omega \Omega^T)^{-1} \Omega \Lambda \quad (57)$$

Then, $\hat{V}_{i+1}, \hat{u}_{i+1}$, and $\hat{d}_{i+1}$ can be obtained by using (47)-(49). Repeat the above steps, the approximate optimal solution of the $H_\infty$ tracking control can be obtained for the tracking problem.

## V. SIMULATION VALIDATION OF THE OFF-POLICY $\delta$-PI ALGORITHM

In this section, a computer simulation is carried out to demonstrate the effectiveness of the off-policy $\delta$-PI tracking Algorithm 3 proposed above. This is a revised version of example 2 in [18]. The nonlinear system dynamics is as follows

$$\begin{bmatrix} \dot{x}_1 \\ \dot{x}_2 \end{bmatrix} = \begin{bmatrix} x_2 \\ -x_1^3 - 0.5x_2 \end{bmatrix} + \begin{bmatrix} 0 \\ 1 \end{bmatrix} u + \begin{bmatrix} 0 \\ 1 \end{bmatrix} d \quad (58)$$

The reference trajectories to be tracked for system states are considered as

$$r_1 = \sin(1.5t) \quad (59)$$
$$r_2 = 1.5\cos(1.5t) \quad (60)$$

which can be described by the following command generator function

$$\begin{bmatrix} \dot{r}_1 \\ \dot{r}_2 \end{bmatrix} = \begin{bmatrix} 0 & 1 \\ -2.25 & 0 \end{bmatrix} \begin{bmatrix} r_1 \\ r_2 \end{bmatrix} \quad (61)$$

with initial condition $r = [1, 0]$. Then, one can get the augmented system

$$\dot{X} = \begin{bmatrix} e_{d2} \\ -(e_{d1} + r_1)^3 - 0.5(e_{d2} + r_2) + 2.25r_1 \\ r_2 \\ -2.25r_1 \end{bmatrix} + \begin{bmatrix} 0 \\ 1 \\ 0 \\ 0 \end{bmatrix} u$$
$$+ \begin{bmatrix} 0 \\ 1 \\ 0 \\ 0 \end{bmatrix} d \quad (62)$$

where $X = [e_{d1}, e_{d2}, r_1, r_2]^T$, $e_{di} = x_i - r_i, i = 1,2$. The performance index is chosen as (8) with $Q = 10\mathbf{I}$, $\mathbf{I}$ is an identity matrix with appropriate dimension, $R = 1, \gamma = 5$, and $\alpha = 0.1$.

The activation function $\rho(X)$ of critical NN is selected as polynomial function, which contains the even powers of state variable of the augmented system up to order four, and the activation functions $\phi(X)$ and $\varphi(X)$ of actor NNs contain the odd powers of state variable of the augmented system up to order three, that is

$$\rho(X) = [X_1^2, X_1 X_2, X_1 X_3, X_1 X_4, X_2^2, X_2 X_3, X_2 X_4,$$
$$X_3^2, X_3 X_4, X_4^2, X_1^3 X_2, X_1^3 X_3, X_1^3 X_4, X_1^2 X_2^2,$$
$$X_1^2 X_2 X_3, X_1^2 X_2 X_4, X_1^2 X_3^2, X_1^2 X_3 X_4, X_1^2 X_4^2, X_1 X_2^3,$$
$$X_1 X_2^2 X_3, X_1 X_2^2 X_4, X_1 X_2 X_3^2, X_1 X_2 X_3 X_4, X_1 X_2 X_4^2,$$
$$X_1 X_3^3, X_1 X_3^2 X_4, X_1 X_3 X_4^2, X_1 X_4^3, X_2^4, X_2^3 X_3, X_2^3 X_4,$$
$$X_2^2 X_3^2, X_2^2 X_3 X_4, X_2^2 X_4^2, X_2 X_3^3, X_2 X_3^2 X_4, X_2 X_3 X_4^2,$$
$$X_2 X_4^3, X_3^4, X_3^3 X_4, X_3^2 X_4^2, X_3 X_4^3, X_4^4]^T \quad (63)$$

$$\phi(X) = \varphi(X) =$$
$$[X_1, X_2, X_3, X_4, X_1^3, X_1^2 X_2, X_1^2 X_3, X_1^2 X_4, X_1 X_2^2,$$
$$X_1 X_2 X_3, X_1 X_2 X_4, X_1 X_3^2, X_1 X_3 X_4, X_1 X_4^2,$$
$$X_2^3, X_2^2 X_3, X_2^2 X_4, X_2 X_3 X_4, X_2 X_4^2, X_3^3, X_3^2 X_4, X_3 X_4^2, X_4^3]^T \quad (64)$$

and $W_c \in \Re^{45}, W_a$ and $W_d \in \Re^{24}$.

During the data collection phase, the initial amount of the extended system state variable is set as $X = [-1, 1, 1, 0]^T$, random signals are leveraged as behavior policies, and sampling time step is $T = 0.1$ sec, then collect $N=1000$ data. The data collection stage is shown in Fig. 1 and Fig. 2.

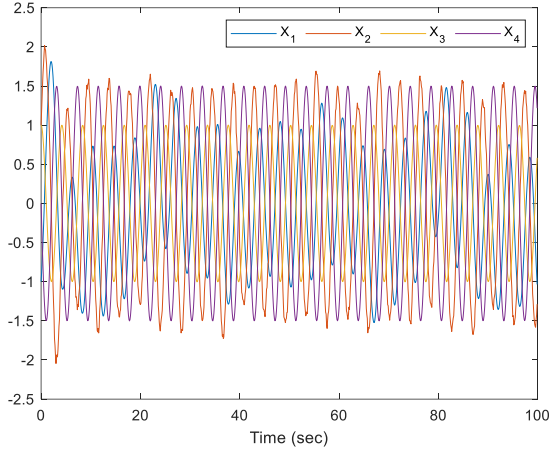

**Fig. 1.** State variable of the augmented system during the data collection phase.

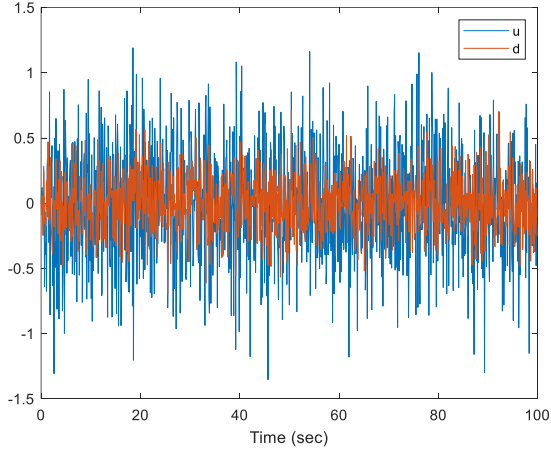

**Fig. 2.** Behavior policies $u$ and $d$ for data sampling.

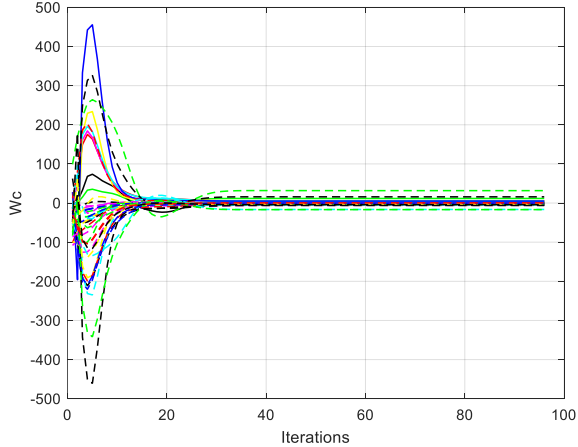

**Fig. 3.** The weight learning process of critic NN.

After collecting the required system data, iterative learning is performed to learn the approximate optimal solution of $H_\infty$ tracking control. Set $W_c^0 = 0$, $W_a^0 = 0$ and $W_d^0 = 0$, and the Newton step-size is taken as $\delta = 0.3$. Reuse the sampled data, solve the weights of the neural network iteratively until $\|W^{i+1} - W^i\| < 10^{-7}$. The learning process is shown in Fig. 3 and Fig. 4. Through 96 iterations, the neural network weights converge to meet the threshold requirements, and the weights of actor neural network are learned.

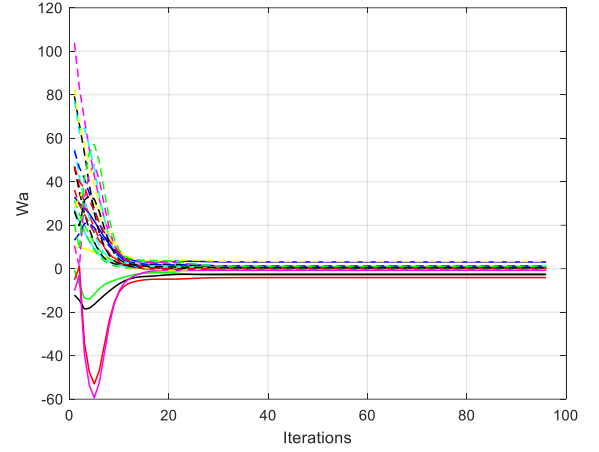

**Fig. 4.** The weight learning process of actor NN.

Leveraging the actor NN weights learned, the $H_\infty$ tracking controller can be obtained through (48). Apply this $H_\infty$ tracking controller to the system to test performance. The initial value of the reference signal and the system state are set as $r_0 = [0, 1.5]^T$ and $x_0 = [0, 1.5]^T$. Set the disturbance signal as

$$d = 1.55\exp(-0.08t)\cos(0.3t) \tag{65}$$

The dynamic response process of the system is shown in Fig. 5-Fig. 7, the tracking error is asymptotically stable when $d \to 0$. Fig. 8 shows the disturbance attenuation level of the $H_\infty$ tracking controller learned by Algorithm 3. As can be seen that the $L_2$-gain condition (5) is met.

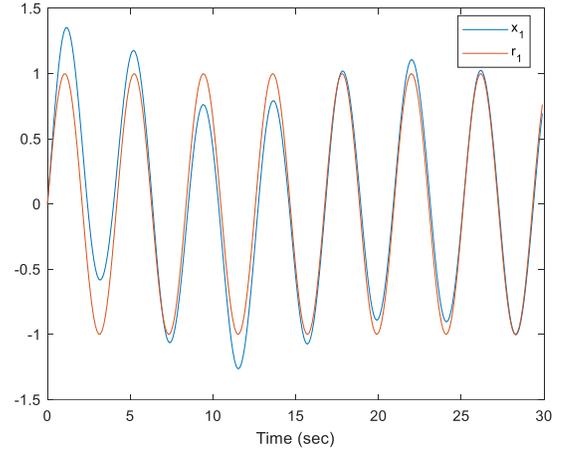

**Fig. 5.** The system state $x_1$ versus the reference signal $r_1$.

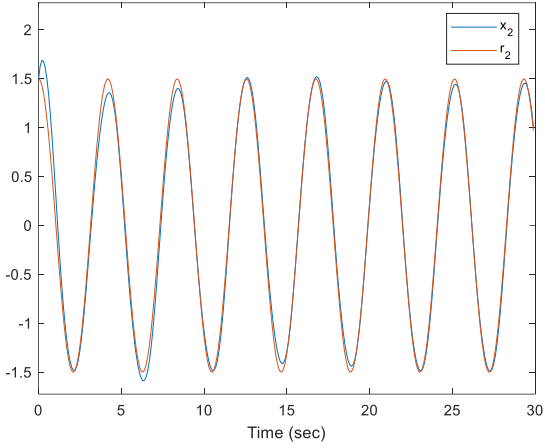

**Fig. 6.** The system state $x_2$ versus the reference signal $r_2$.

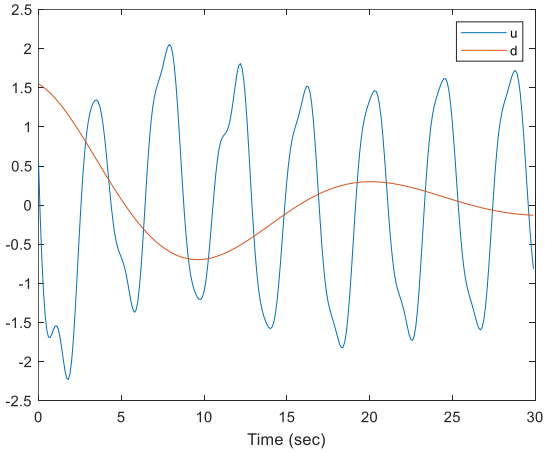

**Fig. 7.** The Control input and disturbance signal.

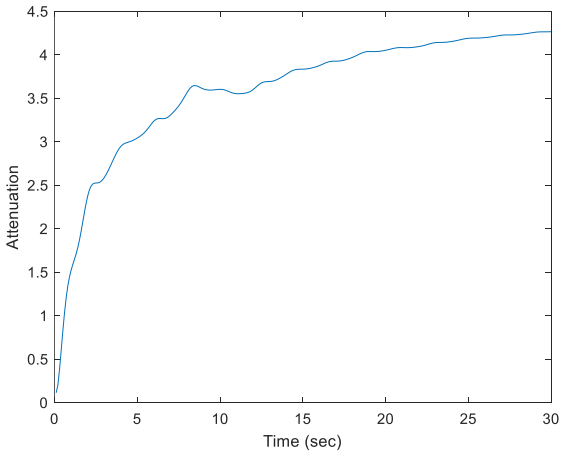

**Fig. 8.** Disturbance attenuation for the $H_\infty$ tracking controller obtained.

## VI. Conclusion

Because damped Newton method enhances the robustness with respect to initial condition and has larger convergence region, a $\delta$-PI algorithm based on damped Newton iteration was developed for $H_\infty$ tracking control of nonlinear continuous-time systems. Firstly, the damped Newton iteration operator equation for solving HJI equation was constructed, then the generalized tracking Bellman equation which is an extension of tracking Bellman equation was derived. Then, by iteratively solving the generalized tracking Bellman equation, on-policy $\delta$-PI algorithm and off-policy $\delta$-PI algorithm were presented, respectively. The off-policy $\delta$-PI algorithm is a model-free algorithm for solving $H_\infty$ tracking control, which can be implemented without making use of any knowledge of the dynamics of the systems. The NN-based implementation scheme for the off-policy $\delta$-PI algorithms was proposed. And finally, the suitability of the off-policy $\delta$-PI algorithm was illustrated with a simulation example. In the Banach space, how to choose the Newton step-size $\delta$ in the $\delta$-PI algorithm to obtain as global as possible convergence and fast convergence will be the focus of the future work.

Acknowledgment

The authors would like to thank the associate editor and the anonymous reviewers for their helpful suggestions and valuable comments.

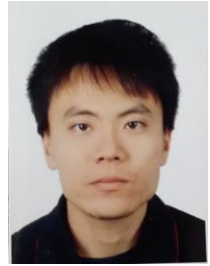

**Qi Wang** was born in Baotou, Inner Mongolia on November 11, 1982. He received the B.E. degree in spacecraft design from Beihang University (BUAA), Beijing, in 2005 and the Ph.D. degree in aircraft design from Chinese Aeronautical Establishment, Beijing, in 2023. He is currently a senior engineer in China Airborne Missile Academy. His research interests are navigation, guidance and control of tactical missile, machine learning, adaptive dynamic programming, and reinforcement learning.

Dr. Wang was a recipient of the Excellent Doctoral Dissertation Award of Chinese Aeronautical Establishment in 2023.